%% file: main.tex
\documentclass[10pt,twocolumn,letterpaper,pagebackref,breaklinks,colorlinks,allcolors=cvprblue]{article}

\usepackage[pagenumbers]{cvpr} 


\definecolor{cvprblue}{rgb}{0.21,0.49,0.74}
\usepackage{hyperref}
\hypersetup{colorlinks,allcolors=cvprblue}
\usepackage{amsmath}
\usepackage{amssymb}
\usepackage{gensymb}
\usepackage{nicefrac}
\usepackage{fontawesome}
\usepackage{pifont}
\usepackage{textcomp}
\usepackage{verbatim}
\usepackage{booktabs}
\usepackage{colortbl}
\usepackage{array}
\usepackage{ragged2e}
\usepackage{tabularx}
\usepackage{makecell}
\usepackage{multirow}
\usepackage{xspace}

\def\paperID{4783} 
\def\confName{CVPR}
\def\confYear{2025}

\title{A Physics-Informed Blur Learning Framework for Imaging Systems}
\author{
    Liqun Chen\textsuperscript{1}\footnote{test} \hfill \quad
    Yuxuan Li\textsuperscript{1} \hfill \quad
    Jun Dai\textsuperscript{1} \hfill \quad   
    Jinwei Gu\textsuperscript{2} \hfill \quad
    Tianfan Xue\textsuperscript{3,1}
    \\[0.4em]
    \textsuperscript{1}Shanghai AI Laboratory \quad
    \textsuperscript{2}NVIDIA \quad
    \textsuperscript{3}The Chinese University of Hong Kong
}

\begin{document}
\twocolumn[{%
\renewcommand\twocolumn[1][]{#1}%
\maketitle
\begin{center}
\vspace{-0em}
    \centering
    \captionsetup{type=figure}
   \includegraphics[width=\linewidth]{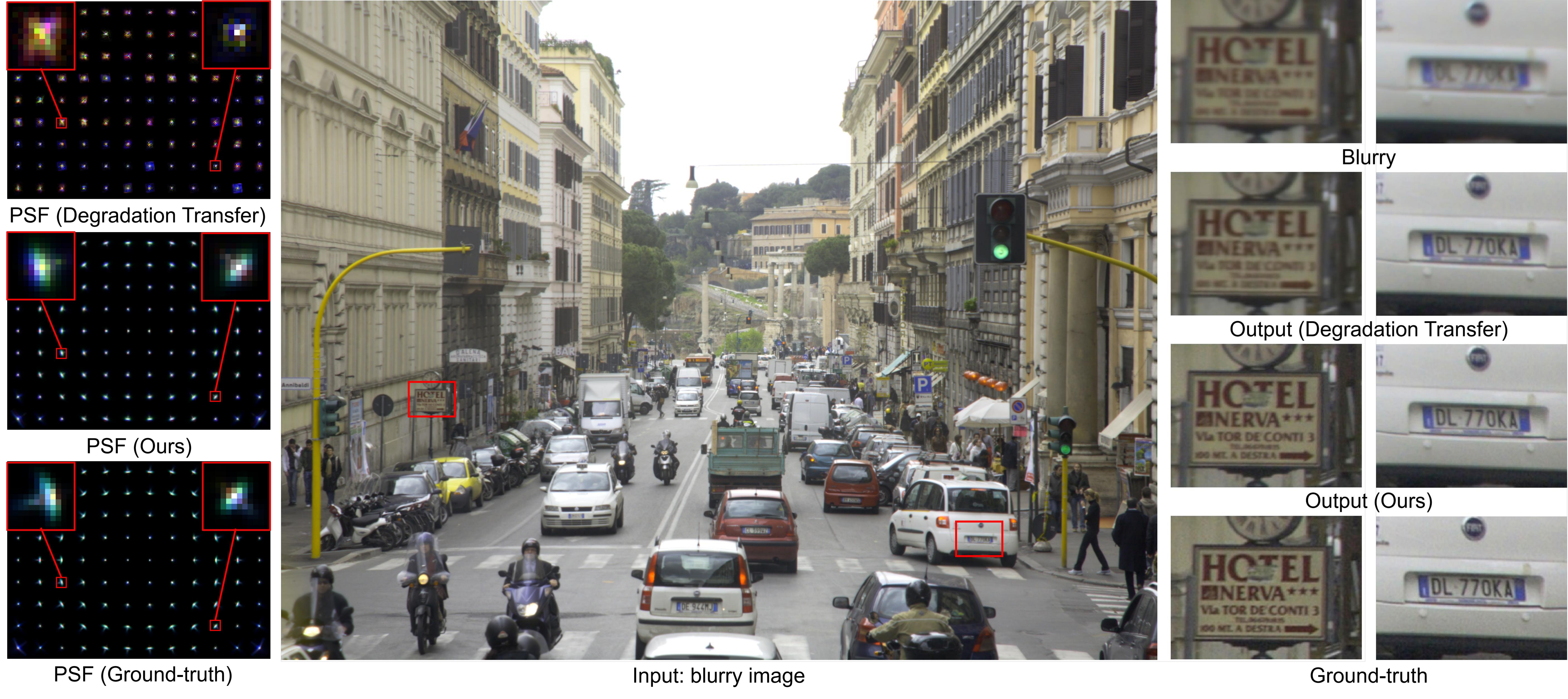}
   \vspace{-2em}
   \captionof{figure}{
    We introduce a point spread function (PSF) estimation framework and demonstrate its effectiveness in deblurring. From left to right: PSF estimated via Degradation Transfer~\cite{chen2021extreme} (state-of-the-art), PSF estimated by our method, and the ground-truth PSF of lens \#63762 from Edmund; a blurry input image synthesized using an image from the FiveK dataset~\cite{fivek} and the ground-truth PSF; the patches from the blurry input image; the patches deblurred by pre-trained Restormers~\cite{zamir2022restormer} using training data generated from PSFs obtained through our method and Degradation Transfer~\cite{chen2021extreme}, respectively; and the corresponding ground truth patches. Our approach outperforms existing state-of-the-art method in both PSF estimation accuracy and deblurring quality.}
   \label{fig:teaser}
   \vspace{-0em}
\end{center}
}]

\input{sec/0_abstract}  
\input{sec/1_intro}

\input{sec/2_related_work}

\input{sec/3_method}

\input{sec/4_experiment}

\input{sec/5_conclusion}

{
    \small
    \bibliographystyle{ieeenat_fullname}
    \bibliography{main}
}

\input{sec/6_suppl}



\end{document}

%% file: sec/0_abstract.tex
\begin{abstract}
Accurate blur estimation is essential for high-performance imaging across various applications. Blur is typically represented by the point spread function (PSF). In this paper, we propose a physics-informed PSF learning framework for imaging systems, consisting of a simple calibration followed by a learning process. Our framework could achieve both high accuracy and universal applicability. Inspired by the Seidel PSF model for representing spatially varying PSF, we identify its limitations in optimization and introduce a novel wavefront-based PSF model accompanied by an optimization strategy, both reducing optimization complexity and improving estimation accuracy. Moreover, our wavefront-based PSF model is independent of lens parameters, eliminate the need for prior knowledge of the lens. To validate our approach, we compare it with recent PSF estimation methods (Degradation Transfer and Fast Two-step) through a deblurring task, where all the estimated PSFs are used to train state-of-the-art deblurring algorithms. Our approach demonstrates improvements in image quality in simulation and also showcases noticeable visual quality improvements on real captured images. Code and models are public.

\end{abstract}

%% file: sec/1_intro.tex
\section{Introduction}
\label{sec:intro}

Imaging systems drive broad applications across numerous fields. However, their practical performance is inherently constrained by spatially nonuniform aberrations. Accurately characterizing it is crucial for achieving high performance in digital photography ~\cite{zhang2019deep,yue2015blind,zamir2022restormer,delbracio2021polyblur,gong2024physics}, industrial inspection~\cite{wu2022integrated}, automotive driving~\cite{tseng2021differentiable}, astronomical observation~\cite{karabal2017deconvolution,guo2024direct} and microscopy~\cite{qiao2024deep,zhao2022sparse}. 

The point spread function (PSF) serves as a mathematical representation of blur. Although accurately modeling the PSF in imaging systems offers significant benefits, achieving both high accuracy and broad applicability remains a significant challenge. Despite the numerous methods proposed for PSF estimation ~\cite{jemec20172d,lin2023learning,liang2021mutual,liaudat2023rethinking,eboli2022fast,chen2021extreme,chen2022computational,qiao2024deep,shih2012image,mosleh2015camera,kee2011modeling}, accurately modeling the PSF requires a detailed characterization of the imaging system, which often involves simulating complex compound lenses~\cite{shih2012image,zhou2024optical,chen2021optical,chen2022computational} based on lens design files. Additionally, many of these models are tailored to specific imaging systems~\cite{zhou2024optical,chen2022computational}, limiting their generalization to other setups. This raises an important question: \emph{Is it possible to achieve universal and accurate PSF estimation through a simple calibration, similar to how camera noise calibration is handled in the industry?}

In this work, we propose a physics-informed PSF learning framework for imaging systems, which consists of a simple calibration step followed by a learning process. Our approach is designed to provide both broad applicability and high accuracy.

We propose a novel wavefront-based PSF model that effectively represents the PSF of imaging systems without prior knowledge of lens parameters, making it \emph{applicable to a wide range of imaging systems}. Additionally, we design a learning scheme targeting the spatial frequency response (SFR) measurement at the image plane. To improve estimation accuracy, we structure the basis of our PSF model so that each basis influences only a single SFR direction, allowing for a more accurate fit to diverse SFR measurements. Using curriculum learning~\cite{bengio2009curriculum}, we progressively learn the PSF outward from center to edge. Our learning scheme accelerates convergence with lower loss, resulting in \emph{high accuracy}.

Our PSF estimation framework achieves superior accuracy, outperforming existing methods, as demonstrated in \cref{fig:teaser}. To validate our approach, we compare it with recent PSF estimation methods (Degradation Transfer~\cite{chen2021extreme} and Fast Two-step~\cite{eboli2022fast}) through a deblurring task, where all estimated PSFs are used to train state-of-the-art deblurring algorithms. Quantitative comparisons on the Flickr2K dataset~\cite{lim2017enhanced} show significant improvements in image quality, as shown in \cref{tab:compare_deblur}. Additionally, the deblurred results on real captured images exhibit noticeable visual quality improvements, as shown in~\cref{fig:comparision}.

%% file: sec/2_related_work.tex
\section{Related Work}
\label{sec:related}

\paragraph{PSF of Imaging Systems}

The PSF of an imaging system is multi-dimensional and complex, arising and accumulating throughout the imaging process.

A typical color imaging pipeline comprises three key components: a lens, a sensor with a color filter array, and an image signal processor (ISP)~\cite{ramanath2005color,brown2019color,delbracio2021mobile}. Each component significantly influences the PSF of the system.

The lens degrades image quality, and manufacturing imperfections can exacerbate this degradation. Optical aberrations such as spherical, coma, astigmatism, field curvature\cite{hopkins1985image,wyant1992basic} cause blurring, with chromatic aberrations leading to color misalignment and fringing~\cite{sasian2012introduction}. Together, these aberrations result in spatially variant degradation that differs across color channels. 

The color filter array employs pixel-level filters to capture color, inevitably causing information loss during raw image capture. The ISP processes this raw data into a final image through operations such as gain adjustment, demosaicing, color correction, white balance, gamma correction, and tone mapping. These nonlinear processes further complicate the characterization of image degradation.

In this work, we treat the PSF of an imaging system as an integrated whole and estimate it directly from the final captured image.

\paragraph{Models for PSF}

PSF modeling approaches fall into three categories~\cite{lin2023learning}: non-parametric, parametric, and optical simulation-based methods.

Non-parametric models represent blur as a 2D distribution, disregarding interpixel relationships within a field and connections across fields. These models sparsely sample spatially variant PSF across the field of view. Consequently, their sparse and independent nature limits their ability to capture the high-dimensional characteristics of PSF within imaging systems.

Parametric models, such as heteroscedastic Gaussian~\cite{delbracio2021polyblur,eboli2022fast} and Efficient Filter Flow~\cite{schuler2012blind}, use a limited set of parameters, which can oversimplify the PSF. More advanced methods, including Zernike polynomials~\cite{niu2022zernike} and Seidel aberrations~\cite{zhou2024revealing}, incorporate wavefront aberrations and diffraction effects. These models establish field-dependent relationships~\cite{gray2012analytic,zhou2024revealing}, enabling dense PSF estimation with minimal measurements. However, the complexity of Zernike polynomials may hinder practical use, whereas Seidel aberrations offer a simpler parameterization for system aberrations.

Optical simulation models rely on detailed lens design parameters to generate PSF through ray-tracing or diffraction propagation~\cite{baker2003mathematical} under various configurations. However, acquiring accurate lens parameters can be challenging due to intellectual property restrictions.

\paragraph{PSF Estimation}
Many techniques have been developed to estimate the PSF~\cite{jemec20172d,lin2023learning,liang2021mutual,liaudat2023rethinking,eboli2022fast,chen2021extreme,chen2022computational,qiao2024deep,shih2012image,mosleh2015camera,kee2011modeling}. Accurately estimating PSF in real-world imaging systems often requires real captures due to factors such as manufacturing errors, variability in assembly and changes in system performance over time. Among these techniques, there is a significant focus on learning-based methods that utilize real captures. 

Among these learning-based methods, one category employs non-parametric PSF models, such as applying a degradation framework to learn PSF with optical geometric priors~\cite{chen2021extreme}. However, this approach lacks guarantees of smooth transitions both within the PSF and across the field of view. To address these smoothness issues, a recent method uses a multi-layer perceptron to provide a continuous PSF representation\cite{lin2023learning}. The primary challenge of this approach lies in the complex alignment needed between blurred and sharp patterns, involving procedures such as homography-based perspective projection, lens distortion correction, and radiometric compensation.

Another category adopts a parametric PSF model, such as using a heteroscedastic Gaussian with parameters estimated from closed-form equations based on image gradients. However, this model can be overly restrictive, particularly for first-entry lenses where the blur may not conform to a Gaussian kernel~\cite{eboli2022fast}.

In summary, employing an accurate parametric PSF model is critical for precise estimation. Furthermore, robust and simplified measurements are preferred for operational efficiency.

%% file: sec/3_method.tex
\section{Proposed Method}

Our work aims to present a practical approach for learning the PSF of an imaging system. We utilize spatial frequency response (SFR) measurement, a technique widely used in the industry.
\begin{figure}[b]
\centering
\vspace{-0.2cm} 
    \includegraphics[width=1\linewidth]{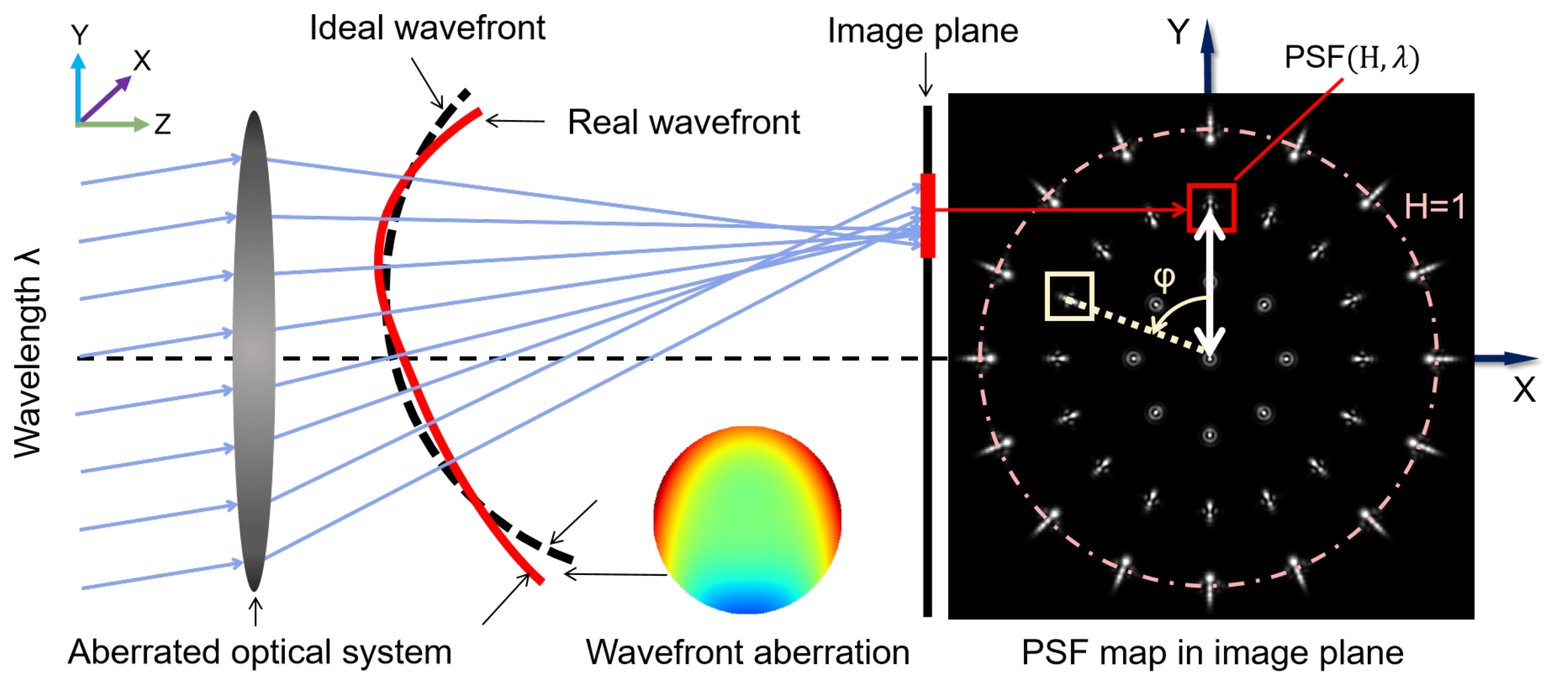}
    \setlength{\abovecaptionskip}{-0.2cm} 
    \caption{Diagram of wavefront aberration and PSF. When light passes through an aberrated optical system, the real wavefront deviates from the ideal, causing defocus in the imaging plane. This deviation, varying with incidence angle and wavelength, creates a spatially varying, symmetric PSF. We focus on the PSF along the +Y axis, where normalized field height $\mathrm{H}$ and wavelength $\lambda$ define $\mbox{PSF}(\mathrm{H}, \lambda)$. Other PSFs are generated by rotating $\mbox{PSF}(\mathrm{H}, \lambda)$ by angle $\phi$ from the +Y axis, with positive $\phi$ indicating clockwise rotation (yellow box).   
 }
    \label{fig:wa_h}
    \vspace{-0.1cm} 
    
\end{figure}

PSF and SFR are interconnected, while the PSF reflects the system's capability to capture fine details, influencing its resolution, the spatial frequency response (SFR) is a key metric for quantifying resolution. The SFR can be derived from the PSF. To simplify the analysis, due to rotational symmetry in optical systems, the PSF is also symmetric (as shown in \cref{fig:wa_h} ). So this study focuses on the PSF along the +Y axis, SFR could be derived from PSF:
\begin{equation}\label{eq:sfr}
\mbox{SFR}(\mathrm{H},\lambda,\phi) = h(\mbox{PSF}(\mathrm{H},\lambda),\phi),
\end{equation}
here, $\lambda$ is wavelength, $\mathrm{H}$ is normalized field height  (shown in ~\cref{fig:wa_h}), and the $\phi$ is the rotation angle from the +Y axis on the image plane, with positive values indicating clockwise rotation (shown in ~\cref{fig:wa_h}), $h$ is the mapping function (see supplementary materials). For a given normalized field height $\mathrm{H}$ and wavelength $\lambda$, the PSF is a 2D distribution, while the SFR relates to directional blur, with the direction specified by the rotation angle $\phi$.

\subsection{Problem Formulation}

\subsubsection{PSF Estimation by Optimization}

The PSF of an imaging system is multi-dimensional, and directly estimating it for different configurations is challenging. A parametric model, such as the Seidel PSF model, can simplify this process.

To understand this model, we start with wavefront aberration, which represents the deviation of the real wavefront from the ideal shape at the exit pupil~\cite{goodman2005introduction}. This deviation leads to defocus on the image plane, resulting in the PSF (see~\cref{fig:wa_h}).  In incoherent imaging systems, the PSF is closely related to the wavefront aberration:  
\begin{equation}\label{eq:psf}
\scalebox{0.95}{$
\mathrm{PSF}(\mathrm{H},\negthinspace\lambda)=\left|\mathcal{F}\negthinspace\left(\negthinspace A( \mathbf{p})\exp\left(\frac{i2\pi W(\mathrm{H},\lambda,\mathbf{p})}{\lambda}\right)\right)\right|^2
$},
\end{equation}
where $W(\mathrm{H},\negthinspace\lambda,\negthinspace \mathbf{p})$ is the wavefront aberration, and $\mathbf{p}$ represents a point in polar coordinates on the pupil plane:

\begin{equation}
\mathbf{p} = \begin{pmatrix} \rho \\ \theta \end{pmatrix},
\end{equation} 
with $\rho \in [0,1]$ as the radial coordinate and $\theta \in [0, 2\pi]$ as the angular coordinate. $A(\mathbf{p})$ is the aperture function, typically known. By further decomposing the wavefront aberration into Seidel basis~\cite{gray2012analytic}, expressed as:
\begin{equation}
\label{eq:seidel}
\scalebox{0.97}{$
W \left( \mathrm{H}, \negthinspace\lambda, \negthinspace\mathbf{p}\right) = \sum \limits_ {\scriptstyle k=0}^\infty\sum\limits_{\scriptstyle l=0}^\infty\sum\limits_ {\scriptstyle m=0}^\infty W_{klm} \mathrm{H}^{k} \rho^{l} \cos^m\left( \frac{\pi}{2}-\theta \right)
$}
\end{equation}
where $ k=2p+m$ and $l=2n+m$ ($p, n, m \negthinspace\in\negthinspace \mathbb{N}$), this decomposition provides a set of Seidel coefficients $ W_{klm}$ (only select around the first 10 items), which theoretically represents a single-channel, spatially varying PSF of an imaging system.

PSF estimation is then framed as learning a set of  Seidel coefficients from observed SFR measurements. This learning is typically achieved through gradient descent optimization. The optimization process aims to adjust the Seidel coefficients to match SFR measurements across the entire image plane.

\subsubsection{Mitigating Gradient Conflicts in Optimization}

\begin{figure}[t]
\centering
\vspace{-0.0cm} 
    \includegraphics[width=1\linewidth]{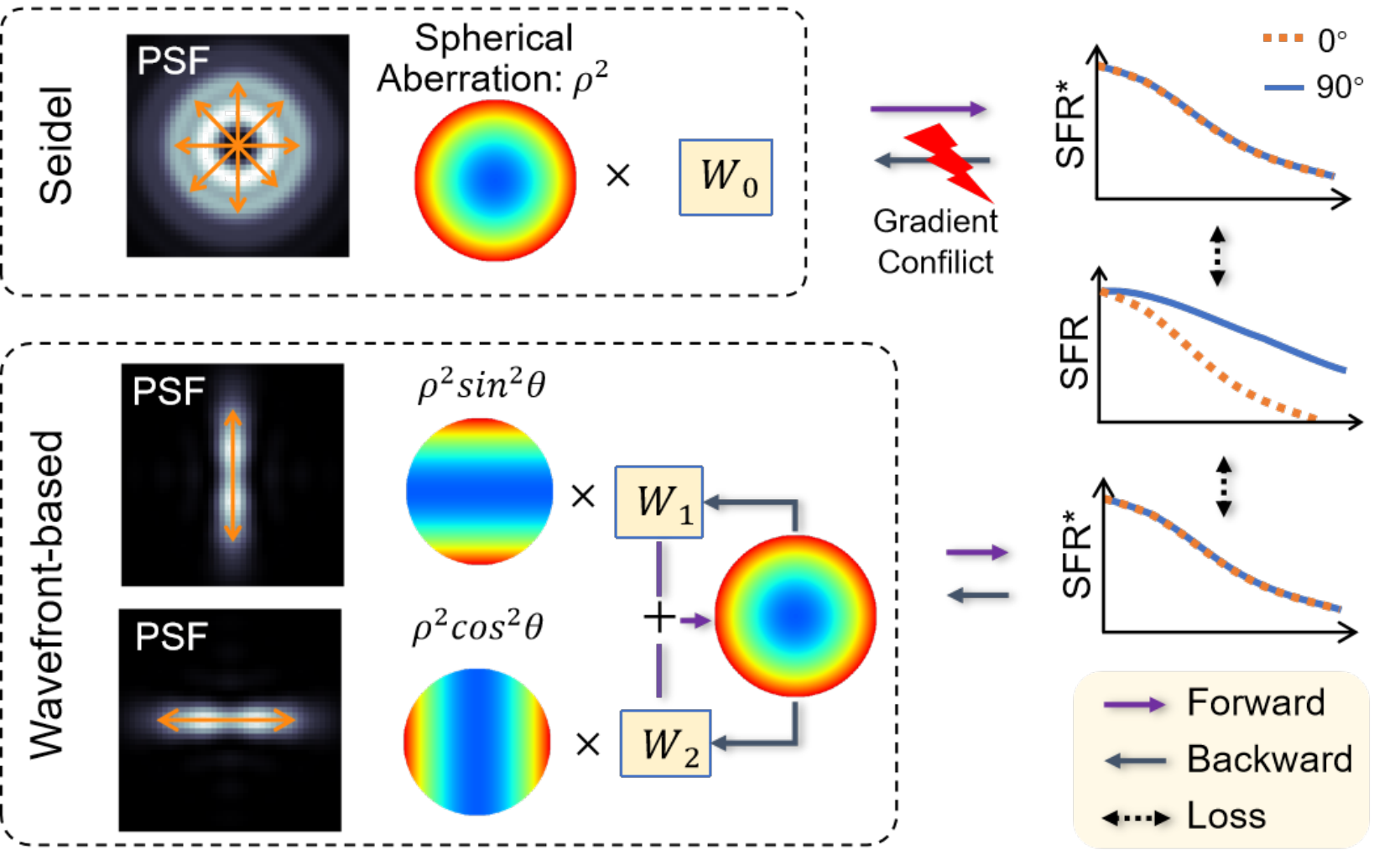}
    \setlength{\abovecaptionskip}{-0.0cm} 
    \caption{An example demonstrating how the proposed wavefront basis mitigates gradient conflicts. \textbf{Top:} In the Seidel PSF model, the spherical aberration basis $\rho^2$ creates a circular PSF shape with $360^\circ$ of blur (orange arrow). This produces identical SFR in both the $0^\circ$ and $90^\circ$ directions. When attempting to optimize the coefficient $W_0$  to match real SFR measurements, which differ between $0^\circ$ and $90^\circ$, gradient conflicts arise. \textbf{Bottom:} In our proposed wavefront basis, each basis affects the SFR in only one direction. This allows the model to independently adjust coefficients $W_1$ and $W_2$ to better match the measured SFR without gradient conflict.
}
    \vspace{-0.0cm} 
    \label{fig:toy}
\end{figure}

However, learning the PSF (or Seidel coefficients) is not trivial for several reasons. Certain Seidel bases, such as spherical aberration ($\rho^2$), simultaneously impact SFR curves across multiple directions, which causes coupling among these directions (\cref{fig:toy}) and hinders accurate fitting to diverse SFR data. Moreover, the inverse problem is inherently ill-posed, particularly due to the exclusion of the phase component in~\cref{eq:psf}. The nonlinearity of transformation as shown in~\cref{eq:psf} further complicates the inversion process. Together, these factors create conflicting gradients during the optimization.

\emph{Gradient conflicts} are frequently discussed in multi-task learning, where the aim is to improve efficiency by sharing model structures across tasks. However, such conflicts can lead to poorer task performance compared to independent learning~\cite{liu2021conflict}. To address this, we build on existing methods for mitigating gradient conflicts and propose refined strategies. 

First, we propose a novel wavefront basis where each basis function influences only one direction of the SFR. The modified expression is:
\begin{align}
\scalebox{0.95}{$
\label{eq:wavefront1}
W(\mathrm{H},\negthinspace\lambda,\negthinspace\mathbf{p}) \!= \!\sum\limits_{\scriptstyle{(p,q,r) \in \mathcal{Q}}} W_{pqr}(\mathrm{H},\negthinspace\lambda) \rho^{p} (\sin \theta)^q (\cos \theta)^{r},
$}
\end{align}
where the set $\mathcal{Q}$ is defined as:
\begin{align}
\label{eq:set_N}
\mathcal{Q} &= \{(2,2,0),(2,0,2), (3,1,0), (3,3,0), (4,2,0), \nonumber\\
&\qquad (4,0,2), (5,1,0), (6,2,0), (6,0,2)\}.
\end{align}

In our modified basis, each term includes either a $\cos \theta$ or $\sin \theta$ component, ensuring it influences the SFR independently along either the vertical or horizontal axis. This approach helps mitigate gradient conflict during optimization, as shown in ~\cref{fig:toy}. For further information about the new basis, please refer to the supplementary materials.

Second, we optimize parameters to match the SFR within narrower field of view. Instead of targeting the SFR across the entire field of view, this approach focuses on smaller, less variable SFR targets, facilitating easier convergence. Although this is a discrete representation, adjusting the optimization step enables control over the PSF output density, allowing for either dense or sparse representations as needed.

Third, we learn the PSF by optimization progressively from the center to the edge~\cite{shi2023deep}. According to aberration theory, only spherical aberration impacts the center of the image plane, while coma and field curvature aberrations gradually appear toward the edges, creating a more complex PSF pattern. Following this progression, we apply curriculum learning~\cite{bengio2009curriculum} to gradually learn PSF from center to edge.

\subsection{Implementation}
\begin{figure*}
\centering
    \includegraphics[width=1\linewidth]{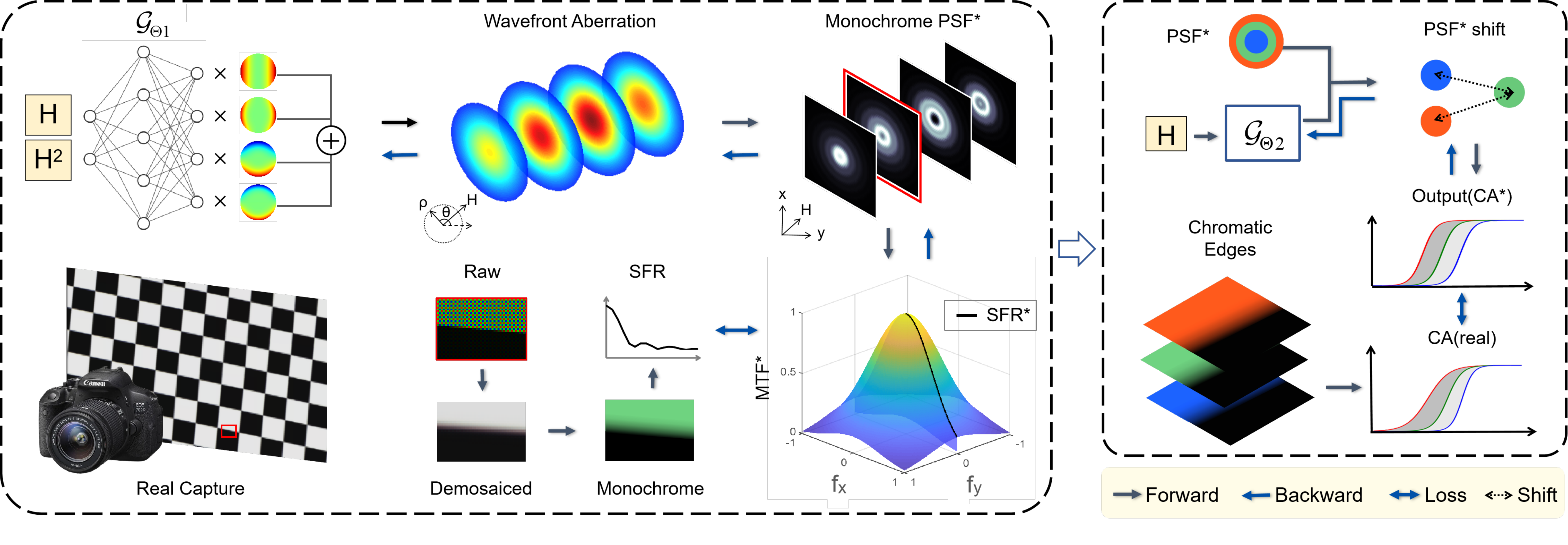}
    \setlength{\abovecaptionskip}{-0.5cm} 
    \caption{Diagram of the proposed two-step PSF estimation framework, the first step involves learning monochromatic aberration per normalized image height $\mathrm{H}$. The network $\mathcal{G}_{\Theta1}$ processes $\mathrm{H}$ and $\mathrm{H}^2$ to output coefficients, generate wavefront aberration and transform it into the $\mbox{PSF}^*$, followed by calculating the modulation transfer function $\mbox{MTF}^*$, resulting in the spatial frequency response ( $\mbox{SFR}^*$) curve. Concurrently, a real $\mbox{SFR}$ curve at the same $\mathrm{H}$ of one color channel is derived from real capture. Discrepancies between these curves guide $\mathcal{G}_{\Theta1}$ to faithfully represent real aberration. The second step focuses on learning PSF shifts across channels. Using $\mathrm{H}$ as input, $\mathcal{G}_{\Theta2}$ calculates shifts, generates shifted PSF, and produces chromatic areas $\mbox{CA}^*$ through a physical process. Real chromatic areas $\mbox{CA}$ data at the same $\mathrm{H}$ are obtained from captures, the disparities between the two data guiding $\mathcal{G}_{\Theta2}$ to output $\mbox{CA}^*$ faithfully representing reality.  These two steps enable the learning of spatial-variant PSF of the whole imaging system.}
    \label{fig:method}
    \vspace{-0.0cm}     
\end{figure*}

\subsubsection{Image Capture}
\label{image capture}

Images are captured in a controlled environment~\cite{ISO12233-2014}, with a checkerboard test chart mounted on a holder and the imaging system positioned on a tripod at a fixed distance. To capture the SFR across the entire field of view in a single setup, the checkerboard is large enough to fill the image plane, and multiple consecutive images are taken. These images are recorded in raw format with the lowest gain, and exposure time is adjusted to prevent overexposure while maximizing the grayscale range. The images are then averaged to reduce noise. Finally, the averaged raw image is converted to a linear RGB format (\cref{fig:method}) for PSF estimation, minimizing the impact of subsequent nonlinear ISP operations that convert linear RGB to sRGB~\cite{brooks2019unprocessing}.

\subsubsection{Two-stage PSF Estimation}

To leverage advanced optimizers and the flexibility of neural networks to enhance the optimization process, we integrate two multi-layer perceptrons (MLPs) into the physical transformations, allowing the MLPs to adjust their neurons to learn the target~\cite{ulyanov2018deep}. As shown in~\cref{fig:method}, our approach follows a two-stage learning strategy. First, we independently estimate the PSF for each color channel. Next, we learn the PSF shifts across channels by analyzing chromatic area differences. Separating the PSF estimation into two subproblems, specifically monochromatic PSF and inter-channel PSF shift, simplifies the optimization process compared to single-stage approach. Here, we refer to the MLPs paired with physical transformations as a surrogate model that represents the PSF of the imaging system.

\paragraph {Monochromatic PSF Estimation}
The MLP $\mathcal{G}_{\Theta_1}$ takes normalized field height $\mathrm{H}$ as input, and outputs coefficients $W_{pqr}^*$. These coefficients are then used to generate the SFR through transformation:
\begin{align}
\mbox{SFR}^*(\mathrm{H},\negthinspace\phi)= h(g(\mathcal{G}_{\Theta_1}(\mathrm{H}) ,\negthinspace\mathrm{H}),\negthinspace\phi),
\end{align}
where $g$ is the mapping function that generates the PSF as defined in~\cref{eq:psf,eq:wavefront1}, and 
$h$ is the mapping function that outputs the SFR as in ~\cref{eq:sfr}, the superscript $*$ denotes the surrogate output, distinguishing it from the ground-truth value. The goal of the surrogate model is to optimize the network parameters to closely match the SFR measurements:
\begin{equation}
\scalebox{0.95}{$
\label{eq:optimization_theta1}
\Theta_1^*(\mathrm{H}) = \mathop{\arg\min}\limits_{\Theta_1} \sum\limits_{\mathrm{H}} \sum\limits_{\phi=0}^{2\pi} \left| \mbox{SFR}^*(\mathrm{H}, \phi) - \mbox{SFR}(\mathrm{H}, \phi) \right|,
$}
\end{equation}
here, in each optimization step, $\mathrm{H}$  is restricted to a smaller region, defined by a narrow field of view, the field of view interval $\Delta\mathrm{H}$ ($\Delta\mathrm{H}\in(0.03,0.1)$). The value $\mathrm{H}$ is gradually increased from 0 to 1 to learn the PSF across the entire image plane.

\paragraph{Cross-Channel PSF Shift Estimation}
In addition to monochromatic aberrations, it is crucial to consider PSF shifts across different color channels, as these shifts can result in color misalignment and fringing, known as chromatic aberration. Building upon previous work~\cite{lluis2012chromatic}, we define the chromatic aberration area (CA) as the region enclosed by the edge gradient line of a blurred black-and-white edge image and the horizontal axis (in pixels). To quantify chromatic aberration, we define the chromatic area difference as:
\begin{equation}\label{eq:deltaca}
\scalebox{0.95}{$
\Delta\text{CA}(\mathrm{H}, \lambda, \phi) = \text{CA}(\mathrm{H}, \lambda, \phi) - \text{CA}(\mathrm{H}, \lambda_{\scalebox{0.6}G}, \phi),
$}
\end{equation}
where $\text{CA}$ is chromatic aberration area, $\Delta\text{CA}$ is chromatic aberration area difference, \(\lambda = \{\lambda_{\scalebox{0.6}{R}}, \lambda_{\scalebox{0.6}{B}}\}\) , and the green channel \(\lambda_{\scalebox{0.6}{G}}\) serving as the reference. 


In practical image capture, chromatic aberration area differences $\Delta\text{CA}$ can be directly inferred from the fringe patterns observed in a captured checkerboard image. However, in a simulated surrogate model, this aberration is influenced by both the 2D distribution of the PSF and the rotation angle \(\phi\), expressed as:
\begin{equation}\label{eq:ca}
\scalebox{0.95}{$
\text{CA}^*(\mathrm{H}, \lambda, \phi)= \mathcal{L}(\text{PSF}_{\scalebox{0.6}{S}}^*(\mathrm{H}, \lambda, \mathbf{x}), \phi),
$}
\end{equation}
where $\mathcal{L}$ is a mapping function (see supplementary materials), and $\text{PSF}_{\scalebox{0.6}{S}}^*$ refers to the shifted $\text{PSF}^*$, which is derived from the monochromatic PSF estimation. To estimate these shifts, we introduce a second MLP \(\mathcal{G}_{\Theta_2}\), which takes the $\text{PSF}^*$ learned by \(\mathcal{G}_{\Theta_1}\), the wavelength  \(\lambda\), and the normalized field height $\mathrm{H}$ as input. It outputs the PSF shifts, which are applied to the PSF as follows:
\begin{align}
\scalebox{0.95}{$
\text{PSF}_{\scalebox{0.6}{S}}^*(\mathrm{H}, \lambda, \mathbf{x}) = \mbox{T}(\mathcal{G}_{\Theta_2}(\mathrm{H}, \lambda), \text{PSF}^*(\mathrm{H}, \lambda, \mathbf{x})),
$}
\end{align}
where $\mbox{T}$  denotes the shift operation, as shown in~\cref{fig:method}. \(\lambda = \{\lambda_{\scalebox{0.6}{R}}, \lambda_{\scalebox{0.6}{B}}\}\), with only the PSF of the red and blue channels being shifted.

The goal is to estimate the PSF shifts between channels to match the chromatic area differences in the measurements. To achieve this, the surrogate model \(\mathcal{G}_{\Theta_2}\) is trained to minimize the difference between the predicted and observed chromatic area differences:
\begin{equation}
\scalebox{0.9}{$
\Theta_2^*(\mathrm{H}, \lambda) = \mathop{\arg\!\min}\limits_{\Theta_2} \sum\limits_{\scriptstyle \mathrm{H}} \sum\limits_{\scriptstyle \phi=0}^{\scriptstyle 2\pi} \left| \Delta\text{CA}^*(\mathrm{H}, \lambda, \phi) - \Delta\text{CA}(\mathrm{H}, \lambda, \phi) \right|,
$}
\end{equation}
where, in each optimization step, $\mathrm{H}$  is restricted to a smaller region and gradually increased from 0 to 1 to learn the PSF shift across the entire image plane, following the same steps and intervals described in~\cref{eq:optimization_theta1}.

Since chromatic area differences arise from both monochromatic aberrations and PSF shifts across channels, we employ a two-stage learning process. The PSF shifts are estimated only after addressing monochromatic aberrations, ensuring a more accurate optimization process.




%% file: sec/4_experiment.tex
\section{Experimental Results}
\label{sec:exp}
\begin{table*}
  \centering
  \resizebox{0.99\textwidth}{!}{
  \renewcommand{\arraystretch}{1.25}
  \begin{tabular}{lc@{\qquad}ccc@{\qquad}ccc}
    \toprule
    \multirow{2}{*}{\raisebox{-\heavyrulewidth}{\centering Comparison Chart}} && \multicolumn{3}{c}{Lens \#63762} & \multicolumn{3}{c}{Lens \#89752} \\
    \cmidrule{3-8}
    && Proposed & Degradation Transfer~\cite{chen2021extreme}& Fast Two-step~\cite{eboli2022fast}& Proposed & Degradation Transfer~\cite{chen2021extreme}& Fast Two-step~\cite{eboli2022fast}\\
    \midrule
    \multirow{2}{*}{\raisebox{-\heavyrulewidth}{H=0 wo/ noise}}
                & PSNR $\uparrow$ &  41.981 $\pm$ 1.132 &  $ 40.822 \pm 1.016$& $ \textbf{42.240} $& 42.073 $\pm$ 1.058& $ 42.128 \pm 1.174$& $ \textbf{43.511} $\\

                & SSIM $\uparrow$   &  0.937 $\pm$ 0.066 & $ 0.939 \pm 0.071 $& $ \textbf{0.943} $& 0.945 $\pm$ 0.069 & $ 0.926\pm 0.073 $& $ \textbf{0.947} $\\

    \midrule
    \multirow{2}{*}{\raisebox{-\heavyrulewidth}{H=0.7 wo/ noise}}
                & PSNR $\uparrow$ &  \textbf{49.185} $\pm$ \textbf{1.242} & $ 46.521 \pm 1.347$ & $ 43.741 $& \textbf{47.811} $\pm$ \textbf{1.115}& $ 45.896 \pm 1.012$& $ 44.171 $\\

                & SSIM $\uparrow$   &  \textbf{0.967} $\pm$ \textbf{0.061}& $ 0.959 \pm 0.079$& $ 0.933 $& \textbf{0.959} $\pm$ \textbf{0.067}& $ 0.951 \pm 0.078$& $ 0.942 $\\
    \midrule
    \multirow{2}{*}{\raisebox{-\heavyrulewidth}{H=1 wo/ noise}}
                & PSNR $\uparrow$ &  \textbf{50.156}$\pm$ \textbf{1.606}& $ 44.920 \pm 1.592$& $44.993 $& \textbf{50.624}$\pm$ \textbf{1.537}& $ 43.872 \pm 1.516$& $44.801 $\\

                & SSIM $\uparrow$   &  \textbf{0.983} $\pm$ \textbf{0.064}& $ 0.966 \pm 0.088$& $ 0.933$& \textbf{0.979} $\pm$ \textbf{0.076}& $ 0.959 \pm 0.081$& $ 0.938$\\
    \midrule
    \multirow{2}{*}{\raisebox{-\heavyrulewidth}{H=0 w/ 1\% noise}}
                & PSNR $\uparrow$ &  \textbf{42.075} $\pm$ \textbf{1.102}& $ 40.629 \pm 1.116$& $ 41.822 $&  \textbf{42.970} $\pm$ \textbf{0.792}& $ 41.053 \pm 1.044$& $ 41.790 $\\

                & SSIM $\uparrow$   &  \textbf{0.949}$\pm$ \textbf{0.065}& $ 0.925 \pm 0.085$& $ 0.937 $& \textbf{0.951}$\pm$ \textbf{0.077}& $ 0.947 \pm 0.080$& $ 0.941 $\\

    \midrule
    \multirow{2}{*}{\raisebox{-\heavyrulewidth}{H=0.7 w/ 1\% noise}}
                & PSNR $\uparrow$ &  \textbf{47.467} $\pm$ \textbf{1.579}& $45.981 \pm 1.483$& $ 44.286 $& \textbf{46.284} $\pm$ \textbf{1.181}& $44.907\pm 1.177$& $ 43.812 $\\

                & SSIM $\uparrow$   &  \textbf{0.960} $\pm$ \textbf{0.071}& $ 0.926 \pm 0.083$& $ 0.950 $& \textbf{0.958} $\pm$ \textbf{0.081}& $ 0.930 \pm 0.087$& $ 0.938 $\\
    \midrule
    \multirow{2}{*}{\raisebox{-\heavyrulewidth}{H=1 w/ 1\% noise}}
                & PSNR $\uparrow$ &  \textbf{49.151} $\pm$ \textbf{1.622}& $ 43.981 \pm 1.629$& $ 44.554$& \textbf{49.803} $\pm$ \textbf{1.643}& $ 43.522 \pm 1.752$& $ 43.604$\\

                & SSIM $\uparrow$   &  \textbf{0.987} $\pm$ \textbf{0.073}& $ 0.936 \pm 0.086$& $ 0.931$& \textbf{0.969} $\pm$ \textbf{0.075}& $ 0.942 \pm 0.082$& $ 0.933$\\
    \bottomrule
  \end{tabular}}
  \caption{Evaluation of PSF accuracy using synthetic checkerboard patterns under different configurations, including variations in relative image height (H) and the presence or absence of noise. The proposed method quantitatively outperforms the Degradation Transfer~\cite{chen2021extreme} and Fast Two-step~\cite{eboli2022fast} methods in terms of PSNR and SSIM. For a fair comparison, all PSFs have been normalized so that the sum of each channel equals one. In most configurations, our method outperforms the other approaches.}
\label{tab:compare_psf}
   \vspace{-2mm}
\end{table*}

We evaluate the proposed method from two perspectives: the accuracy of the estimated PSF in simulations and the deblurring performance in both simulated and real-world scenarios.

\subsection{Dataset, Algorithms and Metrics}

To evaluate deblurring performance, we select three state-of-the-art deep learning-based algorithms: \textbf{MPRNet}\cite{zamir2021multi}, \textbf{Restormer}~\cite{zamir2022restormer}, and \textbf{FFTFormer}~\cite{kong2023efficient}. During the training stage, we use 500 images from the Flickr2K dataset~\cite{lim2017enhanced} to ensure broad applicability across various natural scenes, with the blurred images   synthesized using estimated PSF. During the testing stage, we reserve 100 images from the same dataset, with the blurred images synthesized using ground-truth PSF.

We employ two metric sets to assess performance in simulation and real capture respectively:

\begin{itemize}
    \item \textbf{Full-reference metrics} to evaluate in simulation. We use PSNR and SSIM~\cite{wang2004image} to measure the difference between the output and the ground-truth.
    \item \textbf{Non-reference metrics} are applied for real capture evaluation. We employ MUSIQ~\cite{ke2021musiq} and MANIQA~\cite{yang2022maniqa} to assess the visual quality of the reconstructed images.
\end{itemize}

\subsection{Experiments on Simulation}
We evaluate both the accuracy of the estimated PSF and the deblurring performance by simulation. In this setup, the imaging system uses an IDS camera equipped with onsemi AR1820HS sensor. The imaging lenses are sourced from Edmund (\#63762 or \#89752), and the simulated PSF, generated by \textit{Zemax}\textsuperscript{\textregistered}, serve as the ground-truth. 

To evaluate the accuracy of the estimated PSF, we simulate degraded checkerboard patterns by convolving ground-truth PSF with ideal patterns, followed by estimating the PSF from these degraded patterns. The accuracy of the estimated PSF is compared to the ground truth PSF using PSNR and SSIM metrics. To further assess the robustness of the approach, noise is added to the degraded patterns. For comparison, the following two methods are selected:

\begin{itemize}
\item \textbf{Degradation Transfer}~\cite{chen2021extreme}: A deep linear model incorporating optical geometric priors.
\item \textbf{Fast Two-step}~\cite{eboli2022fast}: An empirical affine model that processes image gradients.
\end{itemize}

An ablation study is then conducted to evaluate the contribution of each component to the overall method.

In evaluating deblurring performance, we account for the ISP pipeline within the camera.

\subsubsection{Accuracy of Estimated PSF}
\label{accuracy of estimate}

\begin{figure}
\centering
\vspace{0.0cm} 
\hspace{-3mm}
    \includegraphics[width=1\linewidth]{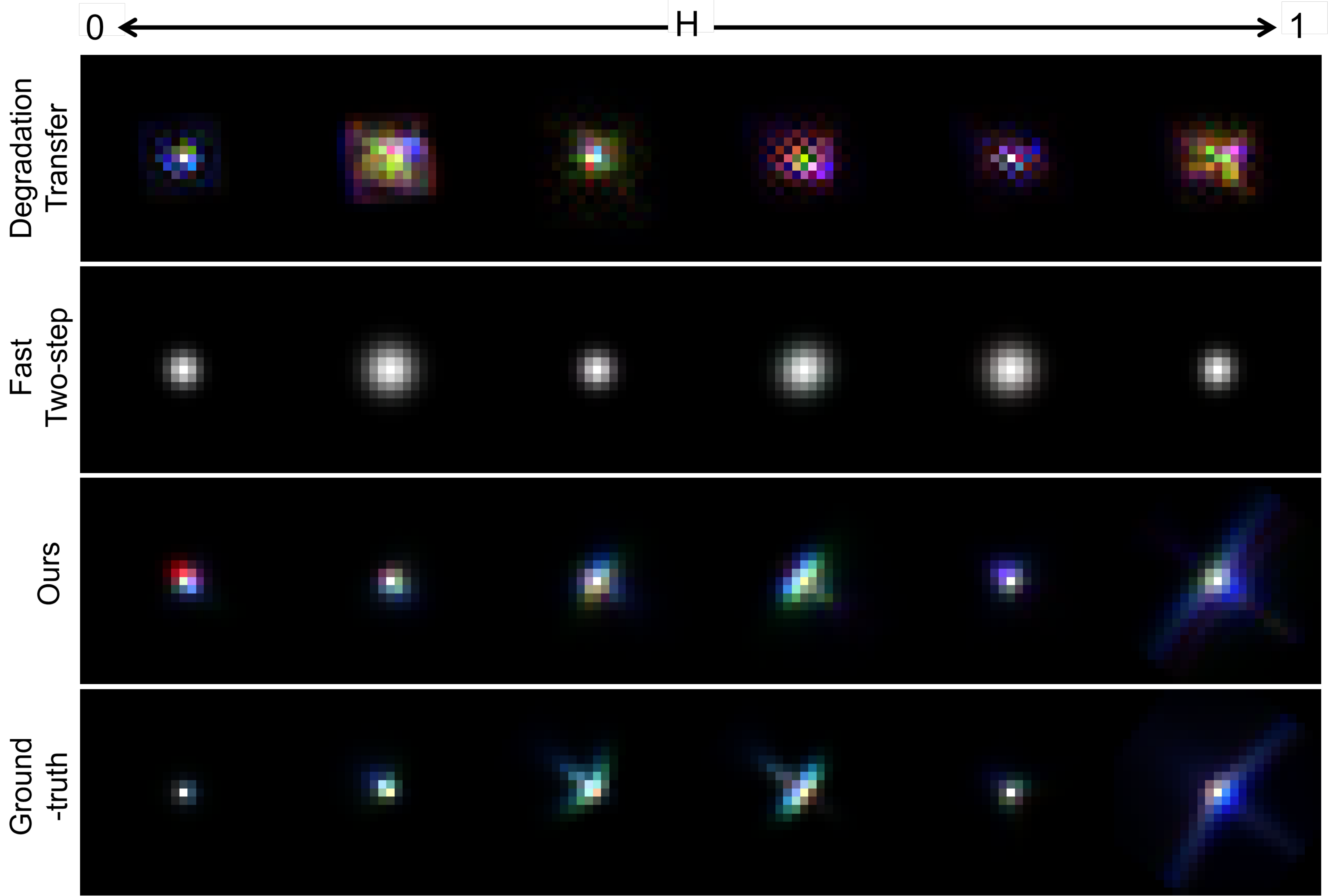}
    \setlength{\abovecaptionskip}{0.3cm} 
    \caption{Estimated PSFs and ground-truth: The PSFs are arranged from left to right by increasing normalized field height $\mathrm{H}$. From top to bottom, the PSF estimates using Degradation Transfer~\cite{chen2021extreme}, Fast Two-step~\cite{eboli2022fast}, and our method, followed by the ground-truth PSF of lens \#63762 from Edmund.}
    \vspace{-0.2cm} 
    \label{fig:psf_compare}
\end{figure}

As shown in~\cref{fig:psf_compare}, the imaging lens is \#63762 from Edmund, and estimated PSF from different methods are listed. PSF is channel-normalized for visualization. Comparatively, our method is closest to the ground-truth.

In traditional lens design, designers typically focus on three normalized field heights: 0, 0.7, and 1~\cite{smith2008modern}, as these provide a representative sampling of the image plane. Following this convention, we selected these normalized field heights for quantitative comparison. We compare two scenarios: one without noise (ideal) and one with noise (realistic) when performing SFR measurements. A 1\% noise level is set for realism, as multiple consecutive checkerboard images can be captured and averaged to reduce noise. As shown in~\cref{tab:compare_psf}, as an optimization method, both our approach and Degradation Transfer~\cite{chen2021extreme} produce variable results, while the Fast Two-step method outputs a consistent result each time.  In most configurations, our method outperforms the other approaches in both scenarios.

\begin{figure*}[t]
\vspace{-0.0cm} 
\centering
    \includegraphics[width=1\linewidth]{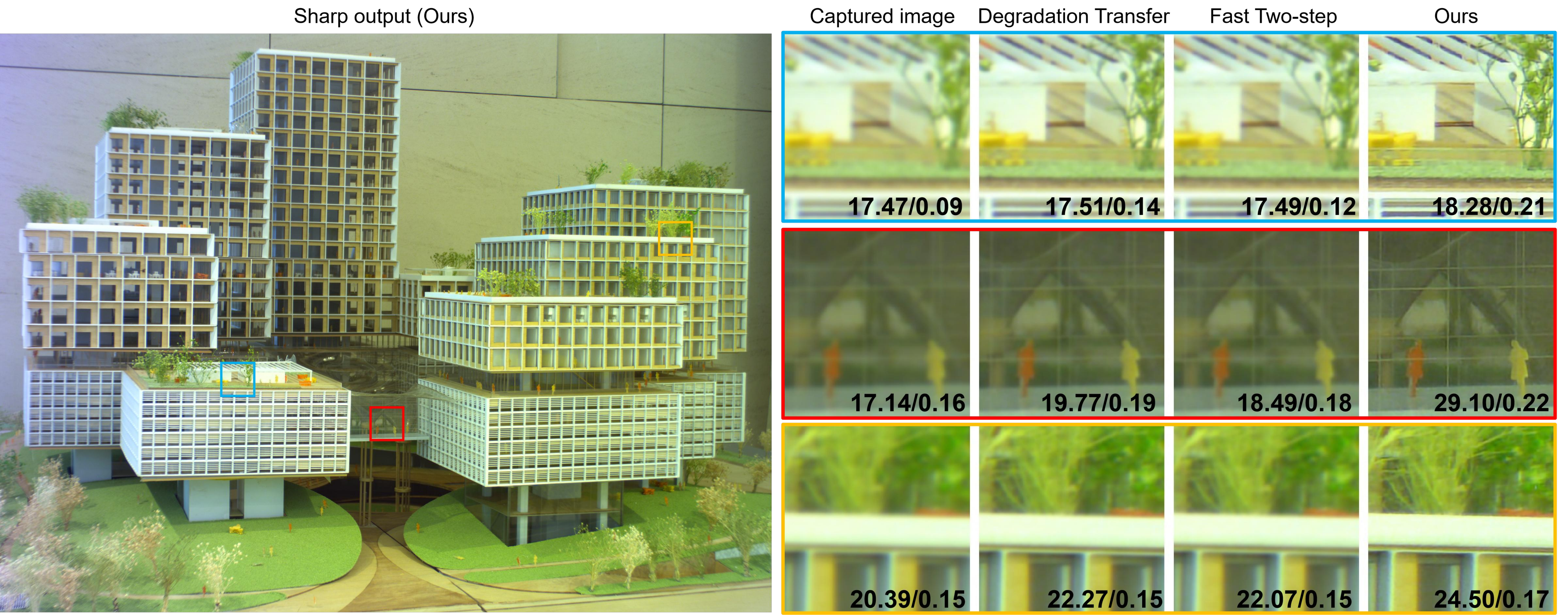}
    \setlength{\abovecaptionskip}{-0.3cm} 
    \caption{Performance comparison with state-of-the-art methods on real captures. From left to right: sharp output image deblurred by the pre-trained Restormer, using training data synthesized from our estimated PSF; real captured image patches from a custom-built imaging system (Edmund Lens: \#63762 and onsemi AR1820HS sensor); deblurred image patches from pre-trained Restormers using data synthesized with estimated PSFs from Degradation Transfer~\cite{chen2021extreme}, Fast Two-step~\cite{eboli2022fast}, and our approach. MUSIQ$\uparrow$ / MANIQA$\uparrow$ scores are shown in the bottom-right corner.}
    \vspace{-0.4cm} 
    \label{fig:comparision}
\end{figure*}

\subsubsection{Ablation Study}

To further evaluate the proposed method, we conduct an ablation study to quantify the impact of various factors on performance. \cref{tab:ablation} presents a comparison of the proposed method with three alternative configurations: (1) without optimization within a narrow field of view, i.e., without small interval optimization; (2) without the proposed wavefront basis in~\cref{eq:wavefront1}, using the Seidel basis instead; and (3) without optimization from center to edge based on curriculum learning, i.e., without curriculum learning. From these comparisons, we conclude that optimization within a narrow field of view, the proposed wavefront basis, and curriculum learning strategy all significantly enhance the estimation accuracy of the spatially variant PSF, particularly for larger fields of view. These design choices are essential for achieving precise results across the entire field of view.

\begin{table}[t]
\vspace{-0mm}
\scriptsize
  \caption{Quantitative assessment (PSNR/SSIM) of each component in the proposed method using the imaging system without noise (Edmund Lens \#63762 and onsemi AR1820HS sensor).}
   \vspace{-2mm}
   \label{tab:ablation}
\renewcommand{\arraystretch}{1.3}
\setlength{\tabcolsep}{5pt}
 \centering
 \resizebox{\linewidth}{!}{
 \begin{tabular}{l|ccc}
        \hline
        &            H = 0&H = 0.7 &H = 1\\
        \hline
        w/o small interval optimization&   42.514/0.934& 42.682/ 0.937& 41.064/ 0.922\\
        w/o  proposed wavefront basis& 42.180/0.931&47.479/ 0.950&  44.579/ 0.954\\
        w/o curriculum learning&  41.562/0.937&48.580/ 0.957&46.023/ 0.955\\
        Proposed&  \textbf{42.643/0.940}&\textbf{49.079}/ \textbf{0.968}& \textbf{49.252}/ \textbf{0.981}\\
        \hline
    \end{tabular}
    }
\vspace{-0mm}
\end{table}


\subsubsection{Deblurring Results}
\label{deblurring results}

\begin{table}[b]
\vspace{1mm}
  \caption{ Quantitative evaluations (PSNR/SSIM) using the imaging system (Edmund Lens \#63762 and onsemi AR1820HS sensor). }
  \vspace{-2mm}
  \hspace{-5mm}
\scriptsize
\renewcommand{\arraystretch}{1.3}
\centering
\resizebox{\linewidth}{!}{
\begin{tabular}{l|ccc}
\hline
\multirow{3}{*}{}           & \multicolumn{3}{c}{Deblurring methods}                            \\ 
                                & MPRNet   & \multicolumn{1}{l}{Restormer} & FFTFormer\\ \hline
Degradation Transfer  w/o noise& 30.546/0.873& 30.691/0.871& 30.534/0.872\\
Fast Two-step w/o noise&            30.217/0.870& 30.340/0.869& 30.335/0.868\\
Ours w/o noise&            \textbf{31.243}/\textbf{0.894}& \textbf{31.506}/\textbf{0.894}& \textbf{31.358}/\textbf{0.891}\\
\hline

Degradation Transfer~\cite{chen2021extreme} w/ noise&            30.326/0.860& 30.417/0.861& 30.308/0.860\\
Fast Two-step~\cite{eboli2022fast} w/ noise&            30.097/0.865& 30.144/0.863& 30.127/0.862\\
Ours w/ noise&           \textbf{31.018}/\textbf{0.889}& \textbf{31.271}/\textbf{0.887}& \textbf{31.145}/\textbf{0.887}\\ \hline
\end{tabular} \label{tab:compare_deblur}
}
\end{table}

Different from the approach in~\cref{accuracy of estimate}, it is crucial to account for the camera pipeline when evaluating deblurring results. To minimize the impact of non-linear operations in the ISP, we assume PSF-induced blur occurs in the linear RGB image.

Thus, we estimate the PSF from a linear RGB checkerboard image. Specifically, a clear checkerboard image is convolved with the ground-truth PSF, followed by mosaicing and demosaicing, to obtain a linear RGB checkerboard image from which we estimate the PSF. We evaluate both noise-free and noisy scenarios during SFR measurement, adding 1\% noise to the blurry checkerboard image in the noisy scenarios. These noisy checkerboard images are then used to estimate the PSF.

The estimated PSF is subsequently used to recover images. We evaluate deblurring performance using deblurring networks~\cite{zamir2021multi, zamir2022restormer, kong2023efficient}. During the training stage, we convert clear images from the Flickr2K dataset~\cite{lim2017enhanced} to linear RGB images through unprocessing~\cite{brooks2019unprocessing}. These images are then convolved with the estimated PSF, followed by color correction and gamma correction to produce blurred images. Both blurred and clear images are fed into the networks for training. In the testing stage, input blurry images are generated using the same process but with the ground-truth PSF.

As shown in~\cref{tab:compare_deblur}, our approach consistently outperforms others in both noise-free and noisy scenarios.

\subsection{Experiments on Real Captures}

We conduct experiments using real captures from the same device used in the simulations (Edmund Lens \#63762 and IDS camera with onsemi AR1820HS sensor).

We capture checkerboard images in the laboratory to estimate the PSF, followed by training Restormer~\cite{zamir2022restormer} (as described in~\cref{deblurring results}). The pre-trained Restormer is subsequently applied to deblur the captured images. For comparison, we follow the same procedure with two other PSF estimation methods~\cite{chen2021extreme, eboli2022fast}.


\subsubsection{Experiment Setup}


We capture checkerboard images in the laboratory using a custom-built device comprising an Edmund Lens \#63762 and an IDS camera (onsemi AR1820HS sensor). The camera is mounted on a tripod, aimed at a checkerboard secured on a card holder, with two angled LED surface lights positioned vertically to provide uniform illumination~\cite{ISO12233-2014}. See \cref{image capture} for further setup details.

\subsubsection{Recovery Comparison}
We estimate the PSF according to the process outlined in~\cref{fig:method}. These estimated PSF are then applied to recover images, followed by an evaluation of the deblurring performance. To reduce cumulative degradation in ISP pipeline, we assume that convolution takes place in the linear RGB domain. Under this assumption, we estimate PSF from linear RGB checkerboard images. To prepare images in the training stage, we first convolve the PSF with linear RGB images generated by unprocessing method~\cite{brooks2019unprocessing},  then apply color correction, gamma correction, and tone mapping to generate blurry sRGB images.

As shown in~\cref{fig:comparision}, a comparison of image patches demonstrates that our method effectively sharpens the image, outperforming others in terms of MUSIQ and MANIQA scores (the higher the better), leading to improved image quality.

%% file: sec/5_conclusion.tex
\section{Conclusion and Discussion}

In this work, we propose a novel physics-informed blur learning framework for imaging systems, which significantly improves the accuracy of PSF estimation and achieves excellent deblurring performance in both simulation and real-world capture scenarios. Importantly, it operates independently of lens parameters, enabling seamless integration into the mass production of various imaging devices without requiring prior knowledge of lens characteristics. While we demonstrate its effectiveness in photography applications, this approach can also provide valuable insights for enhancing image quality in other imaging systems, such as microscopes, industrial inspection systems, and autonomous driving applications.

Our work is only the first step towards PSF modeling and estimation for general imaging systems. Recovering a PSF is inherently ill-posed due to information loss during image formation. In our current work, chromatic aberrations in wide field-of-view images have not been fully corrected (see supplementary materials for examples of failed cases), which will be addressed in the future.

%% file: sec/6_suppl.tex
\clearpage

\appendix
\renewcommand\thefigure{A\arabic{figure}}
\renewcommand\thetable{A\arabic{table}}  
\renewcommand\theequation{A\arabic{equation}}
\setcounter{section}{0}
\setcounter{equation}{0}
\setcounter{table}{0}
\setcounter{figure}{0}

\setcounter{page}{1}
\maketitlesupplementary


\section{Formula and Concept}


\subsection{From PSF to SFR}

The modulation transfer function (MTF) characterizes the relationship between the point spread function (PSF) and the spatial frequency response (SFR). It is defined as:
\begin{equation}\label{eq:sup_mtf}
\mbox{MTF}(\mathrm{H},\lambda,\mathbf{f}) = |\mathcal{F} (\mbox{PSF}(\mathrm{H},\lambda,\mathbf{x}))|,
\end{equation}
where the vector $\mathbf{x} \in \mathbb{R}^2$ is the spatial location on the image plane, $\lambda$ is the wavelength, and $\mathrm{H}$ is normalized field height. The SFR corresponds to a cross-section of the MTF along a specific orientation $\phi$, given by: 
\begin{equation}\label{eq:sup_sfr}
\mbox{SFR}(\mathrm{H},\lambda,\phi) = \mbox{MTF}(\mathrm{H},\lambda,(-sin\phi,cos\phi) \cdot\mathbf{f}),
\end{equation}
where the $\phi$ is the rotation angle from the +Y axis on the image plane, with positive values indicating clockwise rotation, the vector $\mathbf{f} \in \mathbb{R}^2$ corresponds to the frequency components. For simplify, the SFR can be derived from the PSF as~\cref{eq:sup_mtf,eq:sup_sfr}:
\begin{equation}\label{eq:sup_sfr1}
\mbox{SFR}(\mathrm{H},\lambda,\phi) = h(\mbox{PSF}(\mathrm{H},\lambda),\phi),
\end{equation}
where $h$ is a mapping function that converts the PSF to the SFR.

\subsection{From PSF Shift to Chromatic Aberration Area}

Consider an ideal checkerboard pattern with a black-and-white edge at normalized image height $\mathrm{H}$ and angular coordinate \(\phi\), which denotes the rotation angle from the +Y axis on the image plane. Suppose $\text{PSF}_{\scalebox{0.6}{S}}^*(\mathrm{H}, \lambda, \mathbf{x})$ is shifted PSF after $\mathcal{G}_{\Theta_2}$,  this PSF must be rotated by \(\phi\) to align with the edge direction, expressed as:
\begin{equation}\label{eq:rotatepsf}
\text{PSF}_{\scalebox{0.6}{R}}^*(\mathrm{H}, \lambda, \mathbf{x}') = \text{PSF}_{\scalebox{0.6}{S}}^*(\mathrm{H}, \lambda, R(\phi) \mathbf{x}),
\end{equation}
where the new coordinates $\mathbf{x}'$ are:
\begin{equation}\label{eq:newaxis}
\mathbf{x}' = R(\phi) \mathbf{x},
\end{equation}
and the rotation matrix \(R(\phi)\) is given by:
\begin{equation}\label{eq:rotate}
R(\phi) = 
\begin{pmatrix}
\cos\phi & \sin\phi \\
-\sin\phi & \cos\phi
\end{pmatrix}.
\end{equation}
Here, we introduce the edge spread function (ESF)  to establish the relationship between the PSF and chromatic aberration. ESF is derived by:
\begin{equation}\label{eq:sup_esf}
\text{ESF}^*(\mathrm{H}, \lambda, \phi) = \int_{x \leq \alpha} \int_{y} \text{PSF}_{\scalebox{0.6}{R}}^*(\mathrm{H}, \lambda, \mathbf{x}) \, dy \, dx.
\end{equation}
The chromatic aberration area CA is defined as the integral of the ESF curve:
\begin{equation}\label{eq:sup_ca}
\text{CA}^*(H, \lambda, \phi) = \int_{\alpha} \text{ESF}^*(H, \lambda, \phi) \, d\alpha.
\end{equation}
For simplify,  the chromatic aberration area CA can be derived from the PSF:
\begin{equation}\label{eq:sup_ca_L}
\text{CA}^*(\mathrm{H}, \lambda, \phi)= \mathcal{L}(\text{PSF}_{\scalebox{0.6}{S}}^*(\mathrm{H}, \lambda, \mathbf{x}), \phi),
\end{equation}
where $\mathcal{L}$ is a mapping function from PSF shifts to chromatic aberration.

\section{Seidel Basis and Proposed Wavefront Basis}
\label{math}

\begin{table}[h]
\centering
\footnotesize 
\vspace{-0.3cm} 
\renewcommand\arraystretch{1.1}
\begin{tabular}{>{\raggedright\arraybackslash}p{4mm}|>{\centering\arraybackslash}p{17mm}|>{\centering\arraybackslash}p{15mm}|>{\centering\arraybackslash}p{15mm}}
\hline
  & \multirow{2}{*}{Seidel Basis} & \multicolumn{2}{c}{Wavefront Basis} \\ \cline{3-4} 
  &                               & cos & sin\\ \hline
1 & $\rho^2$                      & $\rho^2\cos \theta^2$ & $\rho^2 \sin \theta^2$ \\ \hline
2 & $\rho^3\sin \theta$           & --  & $\rho^3\sin \theta$ \\ \hline
3 & $\rho^3\sin \theta^3$         & --  & $\rho^3\sin \theta^3$ \\ \hline
4 & $\rho^4$                      & $\rho^4\cos \theta^2$ & $\rho^4\sin \theta^2$ \\ \hline
5 & $\rho^5\sin \theta$           & --  & $\rho^5\sin \theta$ \\ \hline
6 & $\rho^6$                      & $\rho^6\cos \theta^2$ & $\rho^6\sin \theta^2$ \\ \hline
\end{tabular}  
\vspace{-0.1cm}
\caption{Decomposition of Seidel basis into proposed wavefront basis.}
\vspace{-0.2cm} 
\label{tab:basis} 
\end{table}

\begin{figure}[h]
\centering
\vspace{-0.0cm} 
\hspace{-3mm}
    \includegraphics[width=0.9\linewidth]{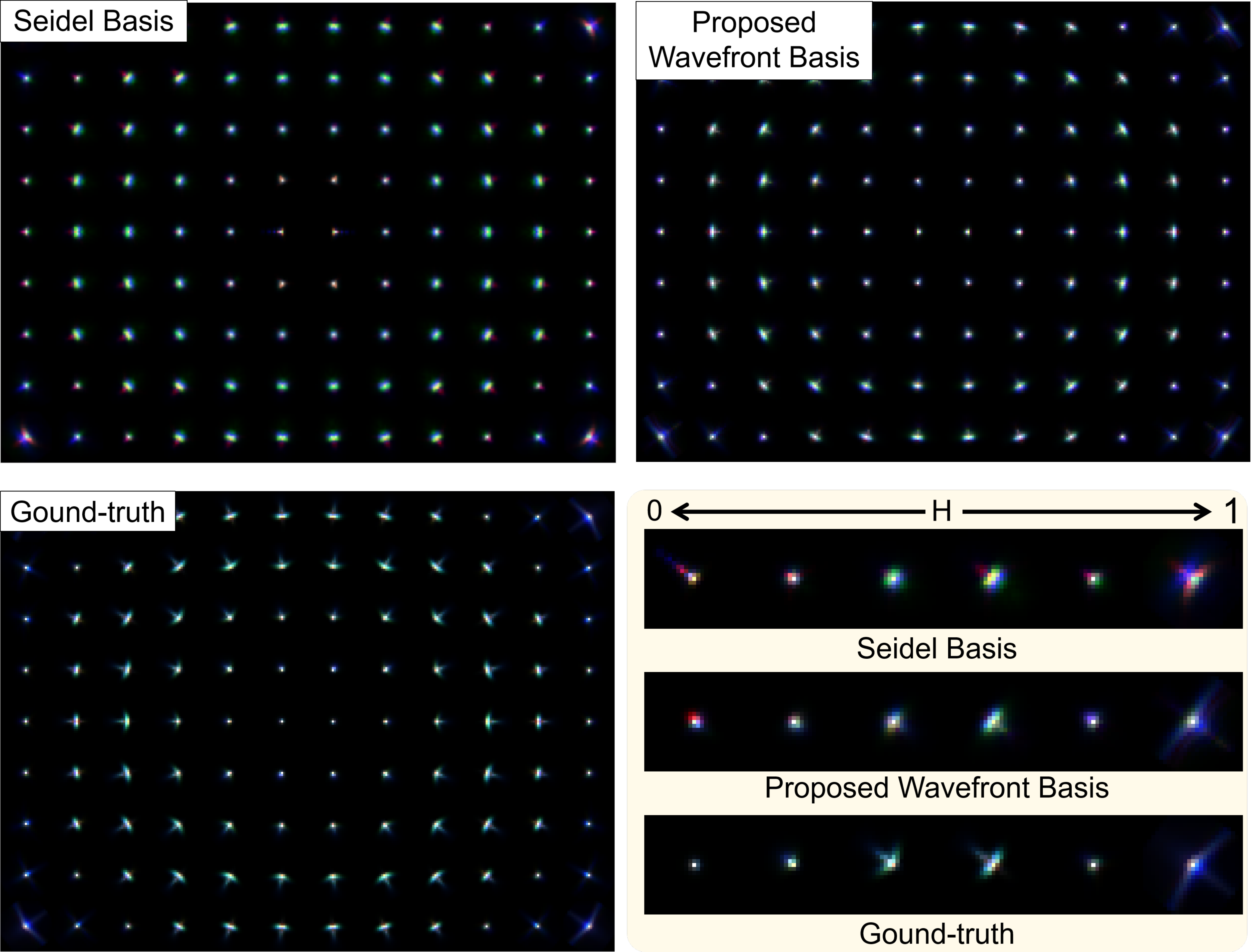}
    \setlength{\abovecaptionskip}{0.3cm} 
    \vspace{-0.1cm}
    \caption{
PSF maps of both the estimated and ground-truth data, with PSFs sampled at evenly spaced intervals along the diagonal of the imaging plane (displayed at the bottom-right).}
    \vspace{-0.3cm} 
    \label{fig:sup_psf}
\end{figure}

As shown in~\cref{tab:basis}, the wavefront basis is obtained by decomposing the Seidel basis. To fully evaluate the proposed wavefront basis, we compare the results optimized with both the Seidel basis and the proposed wavefront basis. As seen in~\cref{fig:sup_psf}, the optimization results using the Seidel basis do not provide a high-accuracy estimation.

\section{Experiments on Real Captures}

\begin{figure*}
\centering
\vspace{-2cm} 
    \includegraphics[width=1\linewidth]{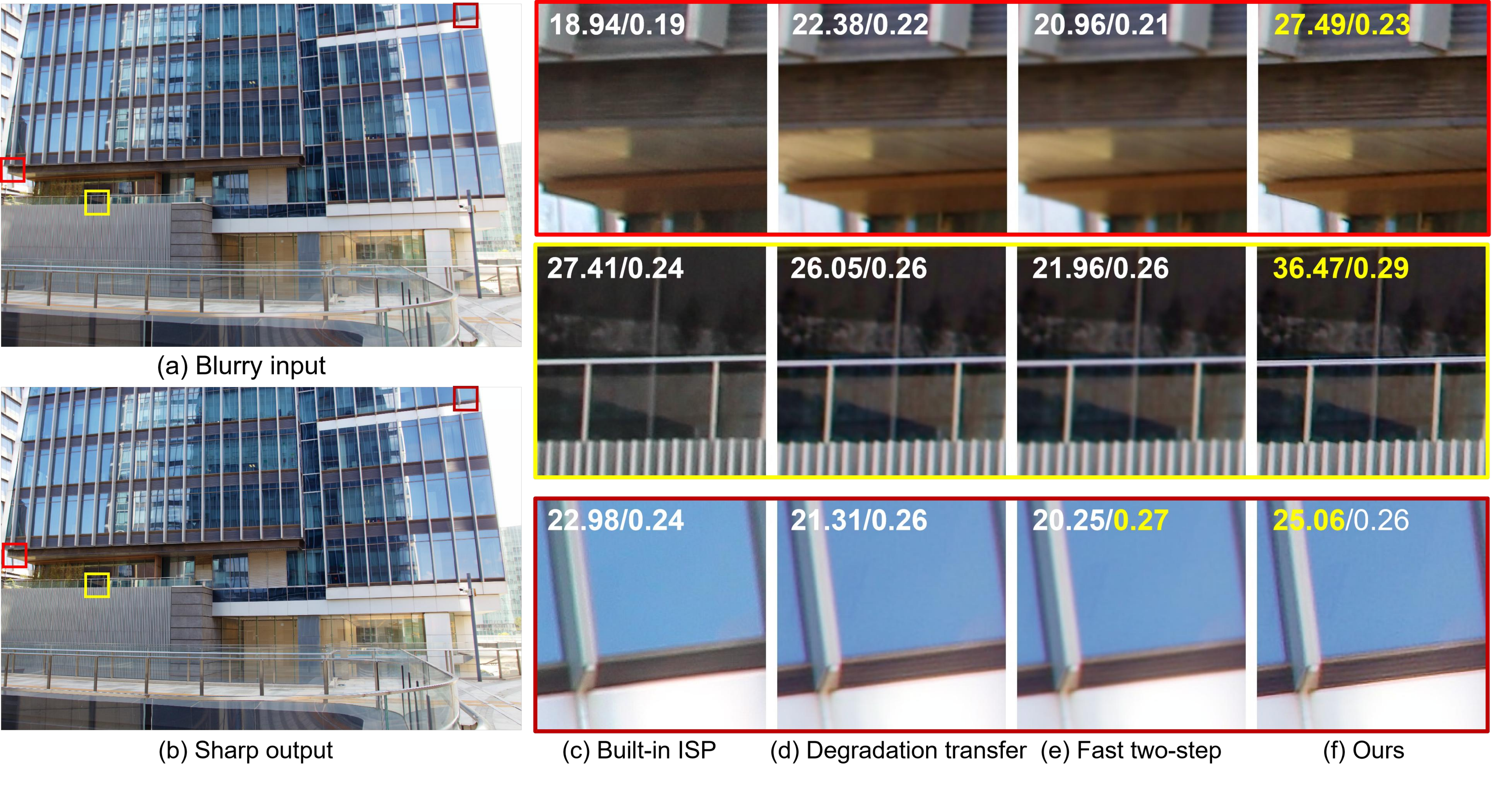}
    \setlength{\abovecaptionskip}{-0.2cm} 
    \caption{Deblurring comparison results. From (a) to (f): blurry input captured with a Canon EOS600D camera, sharp output by our method, images processed by the built-in ISP, and deblurred images processed separately by Degradation Transfer, Fast Two-step, and our method with Restormer. MUSIQ$\uparrow$ / MANIQA$\uparrow$ scores are shown in the top-left corner. As shown, our approach effectively sharpens the image and outperforms the others in terms of MUSIQ and MANIQA scores (higher is better).}
    \vspace{2cm} 
    \label{fig:sup_psf_compare}
\end{figure*}

\begin{figure*}
\centering
\vspace{-2cm} 
    \includegraphics[width=1\linewidth]{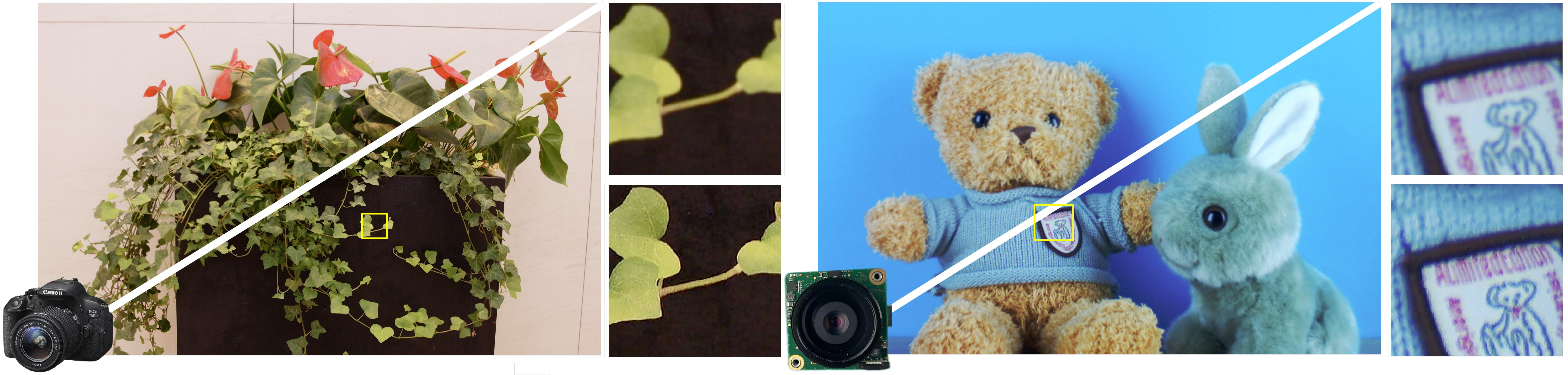}
    \setlength{\abovecaptionskip}{0.1cm} 
    \caption{Validation of the proposed blur learning framework on different devices. Restormer is applied to deblur images trained with the estimated PSF. The captured images are shown in the top-left, and the deblurred images in the bottom-right, with patch comparisons displayed on the right (deblurred patches at the bottom). Left: captured with a custom-built device (Edmund lens \#63762 and Onsemi AR1820HS sensor); right: captured with a Canon EOS600D. As shown, the deblurred image patches reveal more details.}
    \vspace{2cm} 
\end{figure*}

\begin{figure*}
\centering
\vspace{-2cm} 
    \includegraphics[width=1\linewidth]{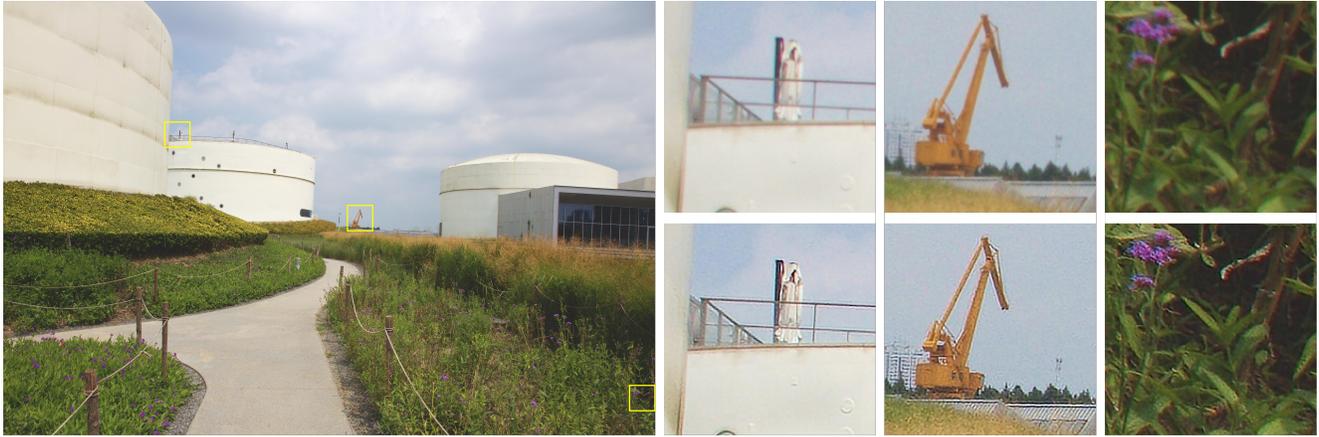}
    \setlength{\abovecaptionskip}{0.1cm} 
    \caption{Deblurring results for an outdoor scene captured with a Canon EOS600D camera. From left to right: sharp output produced by our method, comparison patches (top: captured patches, bottom: patches deblurred by our method using Restormer).}
    \vspace{2cm} 
\end{figure*}

\begin{figure*}
\centering
\vspace{-2cm} 
    \includegraphics[width=1\linewidth]{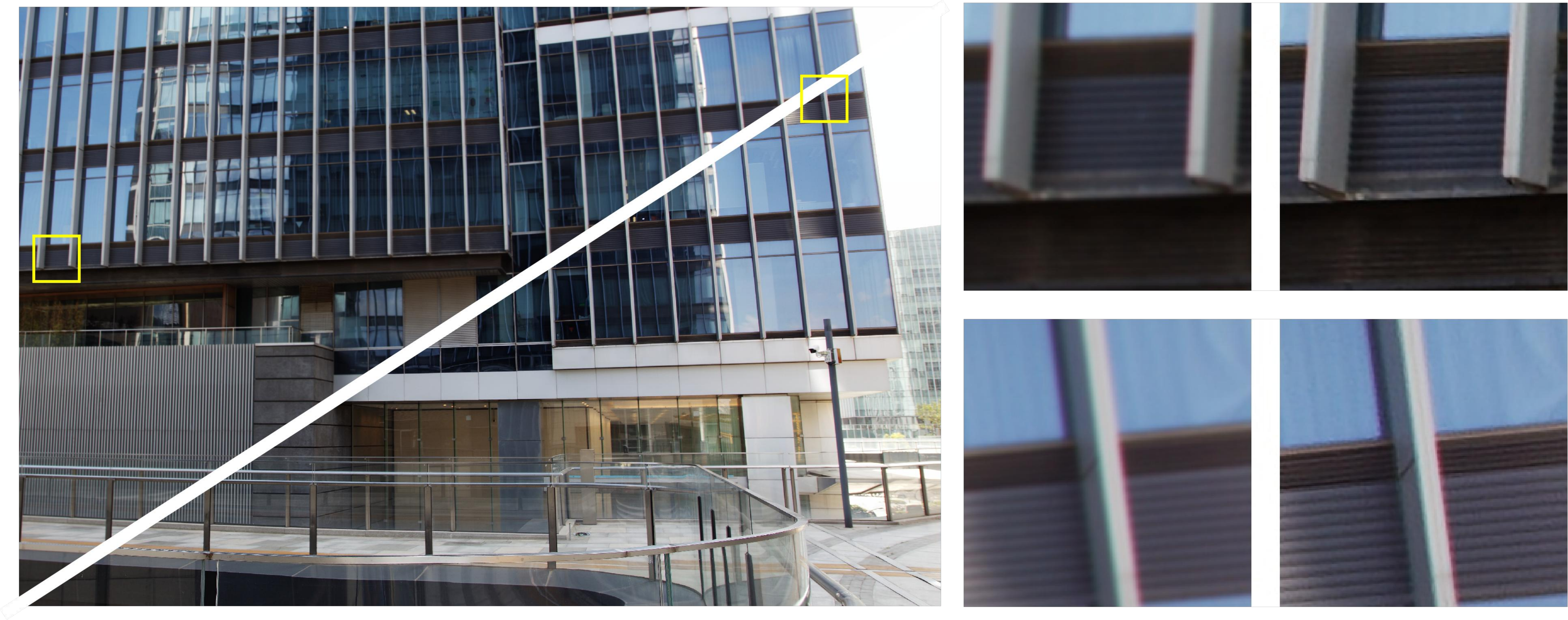}
    \setlength{\abovecaptionskip}{0.1cm} 
    \caption{Failure case, chromatic aberrations in the wide field of view remain partially uncorrected.}
    \vspace{1cm} 
\end{figure*}
